\documentclass{article}

\usepackage[nonatbib, final]{ap_2017}
\usepackage[pdftex]{graphicx}

\usepackage[utf8]{inputenc} % allow utf-8 input
\usepackage[T1]{fontenc}    % use 8-bit T1 fonts
\usepackage{hyperref}       % hyperlinks
\usepackage{url}            % simple URL typesetting
\usepackage{booktabs}       % professional-quality tables
\usepackage{amsfonts}       % blackboard math symbols
\usepackage{nicefrac}       % compact symbols for 1/2, etc.
\usepackage{microtype}      % microtypography
\usepackage{color}

\title{Alquist: The Alexa Prize Socialbot}

\author{
  Jan Pichl \\
  FEE, CTU Prague\\
  Prague, Czech Republic \\
  \texttt{pichljan@fel.cvut.cz} \\
   \And
   Petr Marek \\
   FEE, CTU Prague\\
   Prague, Czech Republic \\
   \texttt{marekp17@fel.cvut.cz} \\
   \And
   Jakub Konr\'ad \\
   FEE, CTU Prague\\
   Prague, Czech Republic \\
   \texttt{konrajak@fel.cvut.cz} \\
   \And
   Martin Matul\'ik \\
   FEE, CTU Prague\\
   Prague, Czech Republic \\
   \texttt{matulma4@fel.cvut.cz} \\
   \And
   Hoang Long Nguyen \\
   FEE, CTU Prague\\
   Prague, Czech Republic \\
   \texttt{nguyeho7@fel.cvut.cz} \\
   \And
   Jan \v{S}ediv\'{y} \\
   CIIRC, CTU Prague \\
   Prague, Czech Republic \\
   \texttt{jan.sedivy@cvut.cz} \\
}

\begin{document}

\maketitle

\begin{abstract}
This paper describes a new open domain dialogue system Alquist developed as part of the Alexa Prize competition for the Amazon Echo line of products. The Alquist dialogue system is designed to conduct a coherent and engaging conversation on popular topics. We are presenting a hybrid system combining several machine learning and rule based approaches. We discuss and describe the Alquist pipeline, data acquisition, and processing, dialogue manager, NLG, knowledge aggregation and hierarchy of sub-dialogs. We present some of the experimental results. 

% The system can be controlled by a text input, or it can be connected to speech recognition services and text to speech services like the Amazon Alexa skill.

%   The abstract paragraph should be indented \nicefrac{1}{2}~inch
%   (3~picas) on both the left- and right-hand margins. Use 10~point
%   type, with a vertical spacing (leading) of 11~points.  The word
%   \textbf{Abstract} must be centred, bold, and in point size 12. Two
%   line spaces precede the abstract. The abstract must be limited to
%   one paragraph.
\end{abstract}

\section{Introduction}

One of the biggest natural language processing (NLP) challenges today is the design of a conversational system. Here we present our contribution the Alquist conversational bot. Alquist is one of the contenders in the Alexa Prize Competition. The primary goal of the competition is to design and implement a bot engaging in a dialogue about popular topics such as latest news, movies, sports, celebrities, and chit-chat. The quality of the bot is determined based on the user rating (users rate the bot on a scale one to five) and the duration of the dialogue. The bot with the highest average rating or if there is a tie, the bot with the longest conversation becomes the winner. Therefore, the goal of the design is to maximize the user experience and the duration.

One of the relatively simple conversational tasks is a goal based dialogue. This task can be compared to form filling. The goal is well defined, for example, a restaurant reservation, air ticket purchase, etc.  The user achieves the goal by answering questions posed by the system. Once the form is filled, the goal is reached, and the dialogue stops. These conversations work reasonably well in Interactive Voice Response (IVR) environments. 
Rule based dialogue managers achieve high accuracy in these tasks, but their development and maintenance are hard and time-consuming. Another disadvantage of the rule based systems is their low flexibility. Rule based system fails when the task is slightly different.

Open dialogues are much more demanding. The dialogue is longer and less constrained. However, it still contains knowledge and information unique to the conversation. It is difficult to design large and unexpected dialogues as rule based systems. Therefore, the recent research focuses on generative, end-to-end models leveraging neural networks \cite{vinyals2015neural}. The goal is to design a system controlling the entire dialogue and to utilize available information about the topic. It is hard to train these systems because we usually do not have enough quality training data. Secondly, we face the problem of extracting and again including entities and entity specific knowledge, as well as utilizing this knowledge during the interaction. 

We have chosen to design a hybrid system combining the end-to-end system and rule based approach. We split the system to smaller rule based topic specific dialogues, such as sports or movies. A good dialogue contains specific information about the conversational topic. Rule based approach and constrained dialogue allow us to work with this information easily. On the other hand, we use the end-to-end dialog implementation once the dialogue diverges from the rule based topics. Naturally, the user experience is lower.

Our way of making the dialogue constrained but keep the impression of an open dialogue is designing a hierarchical system of dialogues. The top dialogue is routing the conversation to topic oriented rule based sub-dialogues. The whole conversation is always going through the top-level dialogue responding to the user in case they change the topic or ask for help. It also allows for some simple commands like ``say it again'' etc. The sub-dialog is composed of even smaller, elementary dialogues taking advantage of their re-usability. These elementary dialogues handle, for example, the yes/no answers or answering a name of a person, location, etc. and can be used in different high-level dialogues throughout the whole system.

With our approach, the dialogues authoring becomes an essential part of a successful conversational system. Good wording helps us to narrow the possible range or users’ replies. Designing such dialogues and still keeping them entertaining is a significant part of a successful design.

Another important part of the conversation is natural language understanding. To lead an engaging conversation about a specific topic, the bot needs to understand it, i.e. it needs to understand user’s intent as well as the information in the uttered sentence. The bot needs to extract the entities and intents to make the conversation successful. We needed a reliable approach for recognizing the intent, the entities, and their relations. Alquist is using different algorithms at different dialogue levels. The selection of intent and entity extraction algorithm was therefore dictated by the state complexity and by the size of available training data. We collected the largest corpora for the top level, which is using a neural network. The lower levels use the information retrieval algorithms representing an utterance by averaged embeddings as will be explained in section \ref{sec:arch}.

\section{Related Work}

The early conversational systems primarily used a pattern matching approach. Eliza \cite{eliza} is one of the most famous chat-bots from that era. Alice \cite{aiml} is another important system which was inspired by Eliza. It uses its own XML based structured language called Artificial Intelligence Markup Language (AIML). Various utterances and their corresponding responses can be stored in this format.

More recent systems can be divided into task oriented and open-domain systems. Furthermore, in \cite{kawahara2009new}, the task oriented systems are divided into those which use a database or statistical text matching as the source of information. They are often used as a tourist guide \cite{mrkvsic2016neural}, reservation assistants \cite{pateras1999mixed} or technical support bots \cite{vinyals2015neural}, because the task oriented systems are suitable for guiding the user using questions to achieve his goal.

Several methods of implementation of the open domain dialogue system have been proposed. The system can be based on natural language understanding \cite{higashinaka2014towards} or the responses can be generated using a recurrent neural network with sequence-to-sequence architecture \cite{vinyals2015neural, bowman2015generating}. Taking advantage of semantic parser outputs such as predicate-argument structure was introduced in \cite{yoshino2011spoken}.

Statistical approaches were proposed to avoid handcrafting the rules for dialogue management (DM). Partially observed Markov decision process (POMDP) \cite{williams2007partially, henderson2015review} has been widely used as a DM since it can handle the uncertainty caused by automatic speech recognition (ASR) or natural language understanding (NLU). The dialogue state tracking challenge \cite{williams2013challenge} has provided several labeled dialogue data as well as the evaluation framework to accelerate research of DM.

The recent work \cite{hcn} shows how to reduce an amount of data required to train end-to-end RNN network using a combined system which uses both RNN and handcrafted domain-specific knowledge.

\section{System overview}

\begin{figure*}[ht]
\begin{center}
\includegraphics[width=\linewidth]{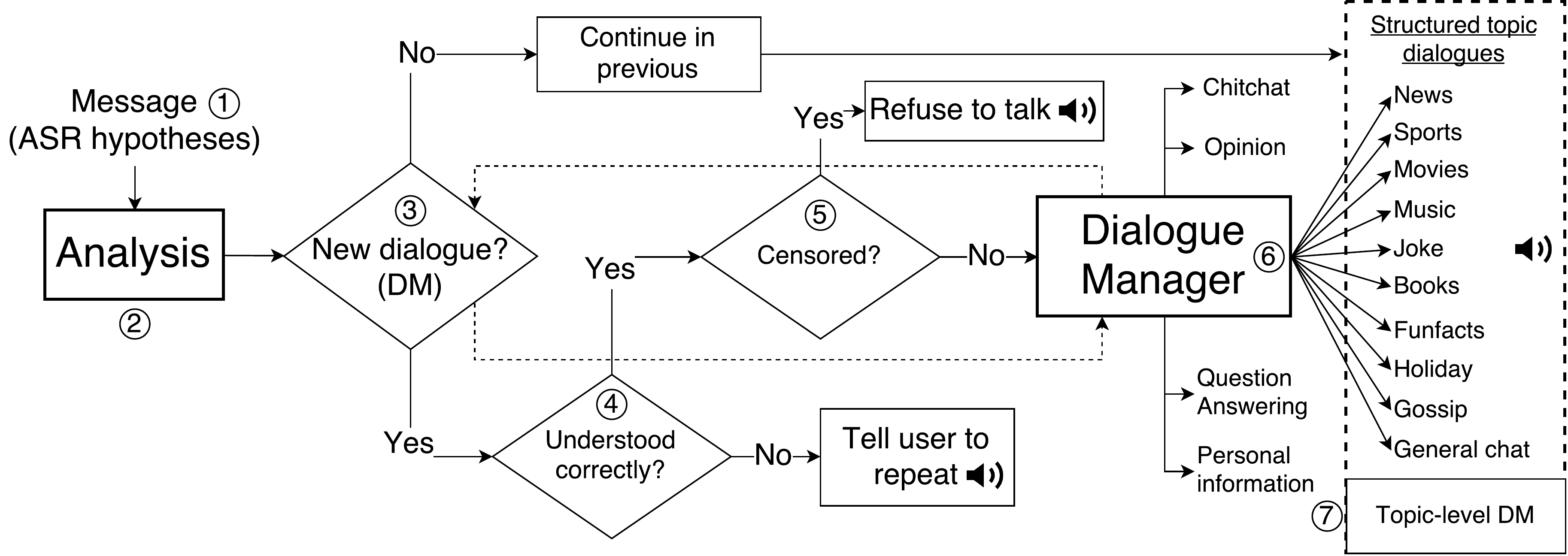}
\caption{The information flow through the system} 
\label{fig:flow}
\end{center}
\end{figure*}

This section describes high-level system architecture and how our system processes the input message. As shown in the Figure \ref{fig:flow}, the system starts with the message received from a Lambda function (1). At the beginning, the message is processed by all of the analysis components (2) described in \ref{subsec:analysis}. After finishing the initial analysis of a message, the pipeline routes the dialog to a Structured Topic Dialogue (3). For example, the user says: ``Let's chat about movies.'' This interrupts the previous dialogue and starts the movie Structured Topic Dialogue. If there is no new dialogue detected, the system tries to continue in previous Structured Topic Dialogue and the topic-level dialogue manager (described later) is called. The idea behind the decision being made before the actual dialogue manager is executed is that the Structured Topic Dialogue can often produce a response despite the low ASR confidence.
% it can overcome the issues when the ASR confidence is low but the Structured Topic Dialogue can produce a response anyway.

If there is no currently running Structured Topic Dialogue, the message activates other components.
The decision whether the bot understood correctly (4) is made based on the token confidence produced by ASR as described in section \ref{subsec:analysis}. Initially, if the bot is not confident about the user's message, it asks the user to repeat it. Secondly, the message is checked for profanity or inappropriate content (5). If the module detects a topic the bot is not comfortable talking about, it kindly asks the user to choose a different topic. This decision is made using a static list of blacklisted keywords and phrases.

Finally, the message is processed by the top-level dialogue manager (6). The DM decides which of the response generation modules (chit-chat, opinion, question answering, personal information or one of the Structured Topic Dialogues) should produce an answer. If the message falls into the structured dialogue category, the dialogue manager also selects the dialogue topic. Each topic has its own rule based topic-level dialogue manager (7). The topic-level DM selects the current state of the dialogue, and the bot produces the corresponding response.

% We have experimented with several approaches in dialogue manager implementation, including TF-IDF, Word embeddings, Logistic regression and Neural networks. We describe these experiments in section \ref{sec:experiments}.

\section{Architecture}
\label{sec:arch}

In this section, we describe the system architecture in detail. Our system takes advantage of the speech recognition and text-to-speech provided by the Amazon Echo, allowing us to work directly with text input and output. The architecture is divided into two categories, Information Aggregation and Alquist Pipeline. Information Aggregation part gathers information from various sources for later use. It is described in section \ref{ia}. Alquist Pipeline is the actual system's core which accepts the message from the user and generates a response. Alquist Pipeline and its components are described in section \ref{pipeline}.

% \footnote{\url{https://developer.amazon.com/alexa-skills-kit}}

\subsection{Information Aggregation} \label{ia}

Since the goal of the system is to maintain an engaging conversation about popular topics such as movies, sports, or politics, external sources of information are required. We take advantage of several types of information sources:
\begin{itemize}
    \item \textbf{Knowledge Bases} include the facts which do not change over time, or whose values change only occasionally. We obtain this type of information from an RDF database Freebase \cite{bollacker2008freebase}.
    
    We also use the Microsoft Concept graph\footnote{\url{https://concept.research.microsoft.com/Home/Download}} to recognize concepts belonging to the individual entities in user's message. We expanded the Concept graph data with additional entries containing movies and games. These entries were obtained from The Movie Database and Internet Game Database respectively.
    
    Additionally, we use The Movie Database, Internet Game Database, Goodreads and Last FM APIs to obtain another movie, game, book, and music related information respectively\footnote{The URLs of the services: \url{www.themoviedb.org}, \url{www.igdb.com}, \url{www.goodreads.com}, \url{www.last.fm}}.
    
    \item \textbf{Regularly updated information} usually depends on the current events or other time-related aspects. As we want to deliver up-to-date content, we regularly save data from several services. We download daily news articles from The Washington Post, and headlines from Today I Learned\footnote{\url{https://www.reddit.com/r/todayilearned/}} subreddit, which we present as fun facts in conversations.  

% \textcolor{red}{\footnote{Such facts can be the area of the USA, the capital city of Germany, \ldots}}
% \footnote{\url{https://www.washingtonpost.com}}
\end{itemize}

\begin{figure*}[ht!]
\begin{center}
\includegraphics[width=\linewidth]{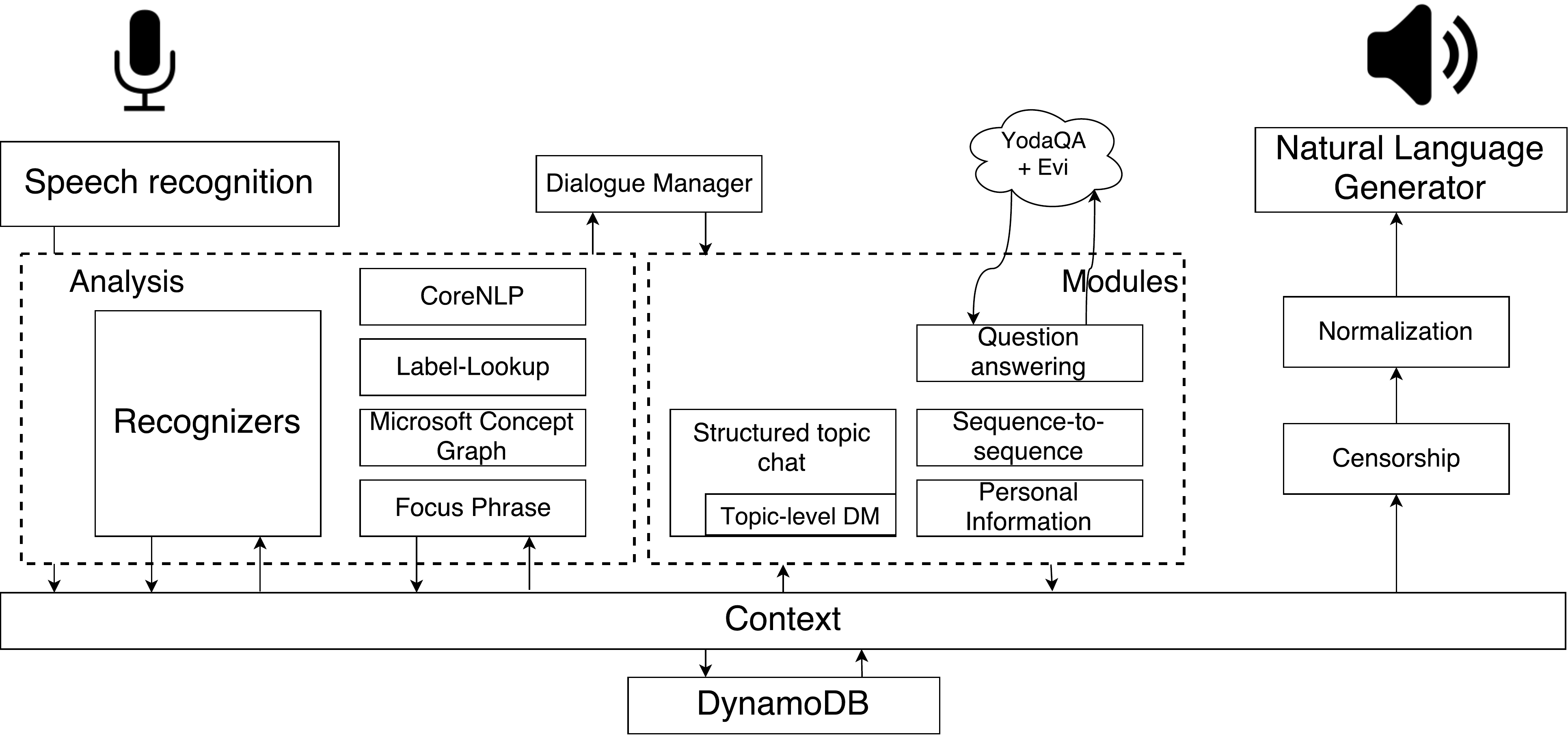}
\caption{The architecture of the Alquist Pipeline. It is separated into logic blocks by the dashed lines. There is central memory called Context used in every part of the pipeline, and its content is regularly stored into a persistent DynamoDB.} 
\label{fig:architecture}
\end{center}
\end{figure*}

\subsection{Alquist pipeline} \label{pipeline}

The Alquist pipeline consists of several interconnected components. It was designed to be highly modular, allowing us to swap individual components and test different approaches quickly. 
% The core system is written in Java, and most of the components are ordinary Java packages. Other components are separated services accessed via REST API allowing us to use different languages (Python) and libraries, especially for machine learning related sub-tasks.

The core Alquist pipeline was inspired by Apache UIMA \cite{ferrucci2004uima}. We adopted several design concepts such as central annotation storage -- we call it context -- where every component stores its results. We describe the context in detail in section \ref{ProcessMemory}.

There are two main types of components in the pipeline, analysis components and response producing components. The purpose of the analysis components is to extract intent and entities from the user's input. The response producing components are either generative or retrieval based neural networks and finite state automata. Additionally, for a specific, small subset of messages, we use various handcrafted responses.\footnote{Handcrafted responses usually answer questions about personal information which require specific lines to be delivered.} All of the response producing components are described in detail in section \ref{subsec:responseprod}

The response producing components are grouped into modules according to the types of the response they produce. We currently employ following modules: Chit-chat, Question Answering, Structured Topic Dialogues, Personal information, and Opinion. These modules are selected by the top-level DM based on information extracted during message analysis.

\subsubsection{Context} \label{ProcessMemory}

% Process memory is the central storage of information about current utterance and the context of the conversation. 
% There is a unique instance of Process Memory for each user's message.
The context contains all the information relevant to the current session. All modules of the pipeline are storing and accessing information in the context. It holds the ASR hypotheses, CoreNLP annotations, recognized entities and eventually the final response. The information that needs to be stored for whole sessions or all following sessions can be marked as a session context or long term context respectively. It can be used, for example, for storing a user's name of user's favorite topics. The information about a topic and last state of ongoing Structured Topic Dialogue is also stored in the context. Additionally, for each topic, the last state and discussed entity are stored in the session context.

The context is regularly (after each response) stored in persistent Dynamo Database. This allows the bot to retrieve the context even if the session was interrupted and to continue with the dialogue.

% Instances of process memory are permanently stored in Amazon Dynamo Database. The data is pulled from the database when the session is interrupted, and the current context needs to be retrieved. Furthermore, the context that is beyond current session is retrieved from the database as well.
% \footnote{\url{https://aws.amazon.com/dynamodb/}}    
\subsubsection{NLP Analysis}
\label{subsec:analysis}

The NLP analysis part of the pipeline uses its components to extract the necessary information from user's message. This information is then used by the dialogue manager and following parts of the pipeline to produce a response. The analysis components can be general or topic-specific. General analysis components are used for all user's messages. These components create tokens, part-of-speech (POS) tags and detect entities from the message. Topic-specific analysis components are used only in certain states of Structured Topic Dialogues. An example of such topic-specific analysis component is specialized entity recognition of book or movie titles.

\paragraph{General analysis components}
We use Stanford CoreNLP \cite{manning2014stanford} for natural language processing (NLP) tasks. The tool provides several annotators. Currently, we use annotators for sentence splitting, tokenization, part of speech tagging, dependency parsing, lemmatisation, sentiment analysis and named entity recognition (NER). Additionally, we take advantage of TrueCase annotation which reconstructs the letter case which was lost by ASR. This is helpful for NER which has a model trained on sentences with original case.

The entity recognition is implemented using two approaches. The first one is Microsoft Concept Graph \cite{concept1, concept2} which provides a list of possible concepts for a given entity (for example ``Frozen'' has following concepts: film, feature, processed food, \ldots). We have a snapshot of the Concept graph\footnote{Snapshot was downloaded in March 2017.} stored in Elasticsearch instance. This allows us to query for a specific entity name and retrieve a concept quickly. The second tool is based on Label-lookup\footnote{\url{https://github.com/brmson/label-lookup}} which is used in question answering system YodaQA \cite{baudivs2015yodaqa}. Label-lookup uses the cross-wikis \cite{Spitkovsky2012ACD} dataset to match strings together with the nearest Wikipedia concept. Our version of the tool has two minor modifications. Firstly we include the Freebase ID in the database and thus eliminate a SPARQL query to DBpedia \cite{lehmann2015dbpedia} to retrieve it. Secondly, we deleted all cross-wiki labels with the Levenshtein distance greater than 3 from the canonical label.
% Label-lookup was initially meant for fuzzy search of concepts in written form, which had a different set of problems than ASR. It was made to be robust against typos, which tend not to happen in speech. We thus have to employ canonical labels and normalize the text. \marginpar{CO JE TO NORMALIZACE TEXTU?}

\paragraph{Topic-specific analysis components}
The group of topic-specific analysis components consists of the recognizers which are suitable for a specific Structured Topic Dialogue. Each topic has a fixed list of keywords. If the keyword is found in the message, it is stored in the context. They are used, for example, for recognition of sports teams, subjects of the news article or movie genre.

We also extract a focus phrase from the user's message as candidate words for entity recognition. This procedure is achieved by a combination of several approaches. The first approach selects all words from the input sentence that are marked as a named entity (NE) by CoreNLP tool. The second approach selects every sequence of consecutive words marked as a noun phrase (NNP) by POS tagger as a focus phrase too. Additionally, a heuristic method based on the dependency parser output selects the lowest noun modification (\texttt{nmod}) of the sentence root and its corresponding adjective modification (\texttt{amod}). We also trained Conditional Random Fields (CRF) sequence labeling using data which were generated using described heuristics and manually corrected. The results of all approaches are combined, and the duplicates are removed.

\paragraph{ASR confidence and profanity filtering}
We compute a speech recognition confidence as follows. We take all ASR hypotheses (up to 5 hypotheses are obtained from the Conversational ASR model provided by Amazon) and their corresponding scores and eliminate some of them based on the following procedure. First, we compute the average of the token probability over the tokens from the sentence. Then, we remove all the sentences with the average smaller than a threshold. The threshold was empirically set to 0.7. The system asks the user to repeat the message if no sentence remains.

The profanity filtering is done simply by searching for banned words from a list. If a word from the list is found in a user's message, the bot refuses to talk about that topic.

% \footnote{\url{https://www.elastic.co/products/elasticsearch}}
\subsubsection{Dialogue manager}
\label{sssec:dm}
There are two types of dialogue managers (DM), top-level DM and topic-level DM. The top-level DM decides which module should be executed (chit-chat, question answering, Structured Topic Dialogue, etc.) and additionally which topic should be executed in Structured Topic Dialogue (sports, movies, etc.). We experimented with several approaches (as described in section \ref{sec:experiments}). The first approach uses similarity of TF-IDF vectors. We convert training examples and user’s input to TF-IDF vector representation. We calculate cosine similarity of user's input to each training example and use the label of the most similar training example as the recognized intent. The second approach uses similarity to sentences represented by a vector calculated as an average of words embeddings in a sentence. The third approach uses a logistic regression. It takes uni- and bi-grams of words from the input message as well as the uni- and bi-grams of Part-of-speech tags.  The last approach uses a convolutional neural network (architecture described in section \ref{sec:experiments}). The input to the neural network is user's message, recognized focus phrase and previous state (module and topic or START if it is a beginning of a conversation). 

In addition to a predicted topic, we implemented a feature which allows the bot to select a topic according to a mentioned entity. Whenever the Structured Topic Dialogue module is recognized, the bot checks if there is an entity matched to some entry in MS Concept Graph. If there is such entity, the bot compares its concept to a predefined list of concepts corresponding to each topic. For each topic, the bot sums up the popularity of the entity given each concept from a list. We select a topic with the highest accumulated popularity count.

Topic-level DM (used in individual Structured Topic Dialogue) is implemented as a rule based system with an extension for switching intents. The rule based DM works as follows. The Structured Topic Dialogue is represented by a state graph. The nodes (states) represent the actions of the system (like a creation of response, saving information into context or accessing the API) and edges are transitions between states. There can be multiple transitions leading from one state. The system has to decide which transition to use in such case. The decision can depend on user's message, recognized entities or data obtained from API for example. If no transition is possible, the system responds by response generated by the generative network.

The system controls the dialogue flow using a state graph. However, we allow the user to take the initiative. We detect an additional set of intents specific to given dialogue, and if one of these intents is detected, the system switches to defined state. This approach allows us to handle a turn-taking conversation with an initiative on both sides. Intent specific to dialogue is detected by embedding similarity based method as described in earlier in this section.

% Neural network based version uses the same graph structure of the dialogue. However, the transition probabilities are predicted by the neural network. The neural network uses a convolutional layer to encode user's message and recurrent layer to preserve the information from the previous sessions.
%  Primarily, we use rule based version of the DM which allows us to gather data for the future network based DM.

\subsubsection{Response production}
\label{subsec:responseprod}
% Response production consists of both rule based and generative approaches.

% We use rule-based approach when our objective is to steer the user towards a more topic focused conversations (e.g. about user's favourite sports team or a popular movie). This approach is used in out Structured Topic Dialogues, Personal information or Opinion module.

% We use generative approach when the user wants to lead more generic, small-talk focused conversation. That is, when we recognize intent that invokes our Chit-chat module.

% If we recognize Question answering intent, we respond with the answer produced by question answering system YodaQA or Evi.

% For these cases we invented a simple YAML syntax, that allows us to randomly generate different utterances from a single line of code and supports swapping in specific entities for template placeholders.

\paragraph{Structured Topic Dialogues}
Structured Topic Dialogues cover the main interaction with the user. 
They are designed to cover the most frequent topics and to provide more in-depth interaction and engaging conversation to the user. The covered topics are books, fun facts, video games, gossip, holidays, jokes, movies, music, news, and sports. The system additionally contains generic dialogue about entities from Freebase, and dialogues providing easier and more fluent interaction with the bot (help dialogue, initial dialogue, exit dialogue, recommendation dialogue). Since our system contains multiple dialogues, we have to decide which one should be executed. This decision is made by the topic-level DM.

The advantage of Structured Topic Dialogues is the ability to maintain and react to the context of the dialogue. Context can be directly encoded into the structure of the state graph. Structured Topic Dialogues can also utilize the context by choosing different transitions in states.

We developed a simple YAML syntax, which can be used to create a response by a random combination of predefined response parts. It also supports swapping in specific entities for placeholders. This allows us to create a graph of the dialogue and lets our system to generate the actual response.
The disadvantage of this approach is that the in-depth  Structured Topic Dialogues is difficult to maintain and test. On the other hand, it allows us to precisely control system's responses.

\paragraph{Personal information and opinions}
The personal information module handles questions about the bot personality (bot favorite color, age, movie preference, \ldots). These messages do not require to be processed as a Structured Topic Dialogue. Instead, a list of handcrafted responses is used to answer a subset of questions. The most similar question is selected using a TF-IDF as described in \ref{sssec:dm}. The opinion related questions (i.e. ``What's your opinion on the latest Super Bowl'') are processed in the same way.

\paragraph{Generative Networks} \label{gn}

The generative neural networks are used for a chit-chat dialogue. Chit-chat is a typically a conversation with no specific topic or any required knowledge about a set of entities. This is for example conversation about user's (bot's) mood, a question about the day, etc. It is triggered when the chit-chat module is recognized or if no other module generates a response.

We use sequence-to-sequence \cite{sutskever2014sequence} network architecture to generate the responses. This framework is commonly used for machine translation but was recently proposed as a conversational model as well \cite{vinyals2015neural}. This model consists of encoder and decoder. The encoder transforms the input sentence with variable length into a fixed size representation, and the decoder decodes the fixed size vector into an output sentence.

We use \texttt{easy\_seq2seq}\footnote{\url{https://github.com/suriyadeepan/easy\_seq2seq}} implementation of the sequence-to-sequence model. The parameters were empirically selected as follows. We set the size of both encoding and decoding vocabulary to 50,000, the number of LSTM cells is 1024 in each layer, and the number of layers is 3. Additionally, we set the batch size to 64 and learning rate to 0.8.

As a training data, we used the discussions from Reddit\footnote{\url{https://www.reddit.com/r/datasets/comments/3bxlg7/i\_have\_every\_
publicly\_available\_reddit\_comment/}}. Various topics from January 2015 to May 2015 were processed. We removed messages with special characters and those longer than 20 words from the dataset. The preprocessing step resulted in the dataset of 3,735,209 message-response pairs, and we created a training split consisting of 3M samples. The rest of the samples ware used as a validation data.

\paragraph{Question answering}

The question answering module is used whenever the user asks a factoid question. These questions are recognized by the module recognizer. The actual process of generating answers is delegated to three systems: YodaQA \cite{baudivs2015yodaqa},  Evi\footnote{\url{www.evi.com}} and a system for answering questions about a news article. For YodaQA, we use version based on \texttt{d/movies} branch. This version uses only a RDF database as an answer source and does not use free text sources. It makes the system significantly faster. Also, we use the new version of Label-lookup which eliminates additional SPARQL queries and speeds up the process. Additionally, the modifications proposed in \cite{pichl} are included. If YodaQA fails to answer, the question is answered by Evi.

The last system is used in the Structured Topic Dialogue about the news. 
This system works in three stages. The first stage simplifies the article's sentences which is done by generating triples from the sentences using \cite{Mausam:2012:OLL:2390948.2391009}, and joining these triples into simpler sentences. The second stage generates questions from the new article using \cite{Heilman2011}. The generated question-answer pairs are stored in a database. Both of these stages happen before the user accesses the article. A question posed by a user is matched to the most similar (based on cosine similarity) question in the database and the answer to it is returned.

% \begin{figure}[ht!]
% \begin{center}
% \resizebox{65mm}{!}{\includegraphics{System-architecture}}
% %\input{obr/dr1a.pstex_t}
% \caption{The description of a figure is of the same style as the description of a table; the figure itself is of the environment \texttt{figure}.
% } 
% \label{fig1}
% \end{center}
% \end{figure}

\section{Experiments}
\label{sec:experiments}

Objective evaluation of the end-to-end dialogue system is a difficult task. Several metrics such as BLEU \cite{papineni2002bleu} were adopted, but they do not reflect the real quality of the dialogue \cite{liu2016not}. We consider the length of the dialogue (either time in seconds between the first and last utterance or a number of utterances) and the diversity of responses as the aspects of the high-quality dialogue. We present the experiment results of experiments on various sub-tasks of the bot as well as the average user rating for a given time period.

% However, we do not have any test data to report the diversity of the answers. We can report the average time and number of responses of the testing users over two weeks of testing. We had 22 users who were asked to test our system. However, no further instructions were provided which is the reason the individual experiences depend on the selected topics of the chat. The time of the average session was 1.94 minutes and the average number of responses was 7.73.
% In this section, we present the experiment results on different sub-task of the bot. 
\paragraph{Intent detection} One of the most important parts of our dialogue system is the intent detection used in dialogue manager. We report the accuracy of four methods we tested to recognize intent in the user's message. Results are shown in the table \ref{accuracy-of-intent-detection}. Tested methods of intent detection are: 
% \marginpar{POPSAT DATASET}
\begin{itemize}
    \item \textbf{TF-IDF:}
    % \footnote{\url{http://scikit-learn.org/stable/modules/generated/sklearn.feature_extraction.text.TfidfVectorizer.html}}
    We use the TF-IDF implementation from the scikit-learn library \cite{scikit-learn}. We found the hyper-parameters of TF-IDF by grid search. They are following, \textbf{analyzer:}~word, \textbf{ngram\_range:}~(1,2), \textbf{max\_df:}~0.9, \textbf{min\_df:}~0.0, \textbf{norm:}~l1, \textbf{smooth\_idf:}~False, \textbf{sublinear\_tf:}~True. The rest of parameters has default values.

    \item \textbf{Embeddings:}
    We use pre-trained GloVe \cite{pennington2014glove} vectors. We convert training examples to vector representation by calculating the average of embeddings of example's words. We normalize vector representations of training examples to unit length. We do the same for the user's input. We calculate cosine similarity of user's input to each training examples and use the label of the most similar training example as the recognized intent. We achieved the best accuracy with the glove.42B.300d variant of vectors.
    
    \item \textbf{Logistic regression:} We use the scikit-learn implementation with default parameters and balanced classes. The input is the vector of word uni- and -bi grams concatenated with the vector of POS uni- and bi-grams. 
    
    \item \textbf{Neural network:} We use GloVe embeddings of dimension $ N=300 $ for input words (both message and focus) and a multi-channel convolution \cite{cnn} with channels of size 1 to 5 for message input and 1 to 3 for focus phrase input. The number of convolutions is set to $ N/2 $, the activation function is $ \tanh $, and max-pooling is over the whole sentence. Since the convolutions of message and focus input are separate layers, the resulting vectors are then concatenated. Additionally, the label predicted for the previous message is concatenated to the vector as well. The concatenated output is then fed into a dense layer with output dimension equal to 300, followed by dropout with rate 0.5 and finally with softmax activated dense layer with output dimension equal to a number of classes.
\end{itemize}

 \begin{table}[h]
   \caption{Accuracy of intent detection}
   \label{accuracy-of-intent-detection}
   \centering
   \begin{tabular}{ll}
     \toprule
     \cmidrule{1-2}
     Method     & Accuracy      \\
     \midrule
     TF-IDF & 0.883      \\
     Embeddings     & 0.897     \\
     Logistic regression     & 0.926         \\
     Neural network     & \textbf{0.927}       \\
     \bottomrule
   \end{tabular}
 \end{table}

 \paragraph{Average rating and time of Structured Topic Dialogues} We selected average user's rating and average time spent as a quality metric for our Structured Topic Dialogues. The average rating is calculated as follows. We collect user's ratings (one to five stars) of the whole conversation. The rating of conversation is assigned to all Structured Topic Dialogues which were used in the conversation. We calculate the average of assigned ratings. Time spent in the dialogue and the number of dialogue turns are measured since the start of Structured Topic Dialogue until a different module is recognized by the DM. The average ratings, times and dialogue turns are presented in the table \ref{average-rating-and-time}.
  \begin{table}[h]
   \caption{Average rating, time and number of dialogue turns of Structured Topic Dialogues}
   \label{average-rating-and-time}
   \centering
   \begin{tabular}{l|lll||l|lll}
     \toprule
     \cmidrule{1-8}
     \textbf{S. T. dialogue} & \textbf{Rating} & \textbf{Time} &  \textbf{Turns} & \textbf{S. T. dialogue} & \textbf{Rating} & \textbf{Time} & \textbf{Turns}\\
     \midrule
     Books & 3.774 & 83 s  &6.151 & Jokes & 3.931 & 50 s&4.131\\
     Fun facts  & \textbf{3.972}  & 64 s & 3.823 & Movies & 3.669 & 69 s&6.199\\
     Games & 3.828  & 93 s & \textbf{6.979}  & Music & 3.823 & 35 s&2.947\\
     General chat & 3.616 & 40 s&2.509 & News & 3.763  & \textbf{102 s}&5.655\\
     Gossip & 3.761 & 66 s&4.315 & Recommendation & 3.575 & 15 s&1.403\\
     Help & 3.509 & 24 s&1.025 & Repeat & 2.429 & 14 s&2.392\\
     Holidays & 3.740 & 26 s&2.495 & Sports & 3.767 & 50 s&4.007 \\
     Initial chat & 3.437 & 44 s&3.959 & Stop & 3.682 & 44 s&2.745\\
     \bottomrule
   \end{tabular}
 \end{table}

\section{Conclusion}

We have described the open domain dialogue system named Alquist. It is focused on engaging dialogue based on current topics and general knowledge. Users can chat about books, games, gossip, holidays, movies, music, news, sport, general knowledge or make simple casual conversation without an additional external knowledge. The system combines machine learning (Sequence-to-Sequence Models) and rule based modules (Structured Topic Dialogues) for response generation.

We evaluated average user rating and average dialogue duration in different phases of the dialogue. The most successful conversations (in both rating and duration) are about the topics which are handled by Structured Topic Dialogues. These dialogues incorporate interesting facts about the corresponding entities and additionally, they suggest a related entity or recommend to switch a topic. This helps keeping the user engaged, unlike the chit-chat and question answering which only reply with a single closed response. These closed replies often lead to an early end of the conversation as the user does not know how to respond.

The rule based topic-level DM is a good solution how to implement a new topic without a corresponding dataset. Additionally, we use the DMs to collect data for each topic which can be used as a training set for a neural network DM. The advantage of the neural network based DM that it can be partially transferred from one topic to another. Rule based system, on the other hand, needs to be rewritten almost from scratch.

\bibliographystyle{iso690}
\bibliography{alquist}

\end{document}